\documentclass{article} % For LaTeX2e
\usepackage{iclr2020_conference,times}

\usepackage{amsmath,amsfonts,bm}

\def\eqref#1{equation~\ref{#1}}
\def\1{\bm{1}}

\DeclareMathAlphabet{\mathsfit}{\encodingdefault}{\sfdefault}{m}{sl}
\SetMathAlphabet{\mathsfit}{bold}{\encodingdefault}{\sfdefault}{bx}{n}

\usepackage{hyperref}
\usepackage{url}

\usepackage{color}
\usepackage{xcolor}
\usepackage{multirow}
\usepackage{graphicx}
\usepackage{enumerate}

\title{Low-Resource Neural Machine Translation: \\ A Benchmark for Five African Languages}

\author{Surafel M. Lakew$^{\dagger +}$, Matteo Negri$^{+}$ \& Marco Turchi$^{+}$ \\
$^{\dagger}$University of Trento, Trento, Italy \\
$^{+}$Fondazione Bruno Kessler, Trento, Italy \\ 
\texttt{$^{\dagger}$\{name.surname\}@unitn.it,$^{+}$\{surname\}@fbk.eu}
}

\newcommand{\mn}[1]{{ \textcolor{black}{#1}}}

\iclrfinalcopy % Uncomment for camera-ready version, but NOT for submission.

\begin{document}

\maketitle

\begin{abstract}
Recent advents in Neural Machine Translation (NMT) have shown improvements in low-resource language (LRL) translation tasks. 
% Back-translation based semi-supervised learning (SS-NMT), multilingual NMT (M-NMT) and transfer-learning (TL) are among the prominent advances.
In this work, we benchmark NMT between English and five \mn{African} LRL pairs ({\bf S}wahili, {\bf A}mharic, {\bf T}igrigna, {\bf O}romo, {\bf S}omali [SATOS]).
We collected the available resources on the SATOS languages to evaluate the current state of NMT for LRLs.
Our evaluation, comparing a baseline single language pair NMT model against semi-supervised learning, transfer learning, and multilingual modeling, shows significant performance improvements both in the En $\rightarrow$ LRL and LRL $\rightarrow$ En directions. In terms of averaged BLEU score, the multilingual approach shows the largest gains, up to +$5$ points, in six out of ten translation directions. To demonstrate the generalization capability of each model, we also report results on multi-domain test sets.
We release the standardized experimental data and the test sets for future works addressing the challenges of NMT in under-resourced settings, in particular for the SATOS languages.\footnote{\mn{Data, models and scripts can be accessed at \url{https://github.com/surafelml/Afro-NMT}}}
\end{abstract}

\section{Introduction}\label{sec:intro}
\mn{Since the introduction of}
recurrent neural networks~\citep{sutskever2014sequence,cho2014learningGRU,bahdanau2014neural} and recently with the 
%transformer 
\mn{wide use of the Transformer}
model~\citep{vaswani2017attention}, NMT has shown increasingly better translation quality
%and fluency~
\citep{bentivogli:CSL2018}.
Despite this progress,
%the limitations of NMT have been repeatedly verified when the training data is scarce~\citep{koehn2017six}. Most
\mn{the NMT paradigm is data demanding and still shows  limitations when models
%have to be 
are trained on small parallel corpora \citep{koehn2017six}. Unfortunately,} 
most of the 
%+$7,000$~\citep{campbell2008ethnologue} languages and language varieties that are currently spoken around the world fall under the this low-resource condition.
\mn{7,000+ languages and language varieties currently spoken around the world fall under this low-resource condition. For these languages, the absence of usable NMT models, creates a barrier in the increasingly digitized social, economic and political dimensions.} 
%Hence, building effective NMT models with a limited amount of training examples is a primary challenge in MT research.
\mn{To overcome this barrier, building effective NMT models 
%capable to translate from/into under-resourced languages by learning from 
from small-sized training sets has become a primary challenge in MT research.}

Recent progress, however, has shown the possibility of learning better models even in 
%a resource-scarce 
\mn{this challenging scenario.}
%scenario, these include:
\mn{Promising approaches rely on}
 semi-supervised learning~\citep{sennrich15:subword}, transfer-learning~\citep{zoph16:tf} \mn{and} multilingual NMT \mn{solutions}~\citep{johnson2016google,ha2016toward}.
%approaches.
% However, because of resource availability and the research focus is on previously established benchmarks, the language and geographical coverage of NMT is yet to reach new heights. 
\mn{However, due to resource availability issues and to the
priority given to well-established evaluation benchmarks, the language and geographical coverage of NMT is yet to reach new heights.} 
This calls for further investigation of the current state of NMT using new languages, especially at a time where an effective and equitable exchange of information across borders and cultures is the most important necessity for many.

%Hence, the objectives of this work are:  {\it i}) to benchmark and expand the current boundary of NMT into five more languages that are predominately used in the East African region and never extensively studied using NMT approaches, namely: {\it Swahili, Amharic, Tigrigna, Oromo, and Somali},
\mn{Therefore, the objectives of this work are: {\it i}) to benchmark and expand the current boundary of NMT into five prominent languages used in the East African region (i.e. \textit{Swahili}, \textit{Amharic}, \textit{Tigrigna}, \textit{Oromo}, and \textit{Somali}), which have not been  extensively studied yet within the NMT paradigm,}
%,\footnote{Language and data: see details in Table~\ref{tab:lang-stat}}
% {\it ii}) to investigate the strength and weakness of NMT approaches towards LRL and define open problems, 
\mn{{\it ii}) to investigate  strengths and weakness of NMT applied to LRLs and define open problems, }
% {\it iii}) to release a standardized training data for the five LRL and multi-domain test sets for up to $20$ directions to encourage future research in zero-shot and unsupervised translation between the LRL languages.
\mn{{\it iii}) to release a standardized training dataset for the five LRL, as well as multi-domain test sets for up to $20$ directions, to encourage future research in zero-shot and unsupervised translation between LRLs.}
%In the following sections, we begin with a summary of NMT and the three approaches we considered for LRL NMT modeling. Then, we provide our experimental settings, data statistics, and models. Before our conclusion, we discuss the results and highlight the open challenges for the research community.

% %
% % NMT Approaches for LRL - see appendix. 
% %

\section{Experimental Settings}
\subsection{Dataset and Preprocessing}
For the five languages aligned to English, we collect all available parallel data from the Opus corpus~\citep{tiedemann2012parallel}, including  JW300~\citep{agic-vulic-2019-jw300}, Bible~\citep{christodouloupoulos2015massively}, Tanzil, and Ted talks~\citep{cettolo2012wit3}. 
For pre-training a massive multilingual model for 
%the TL experiments,
\mn{our transfer learning (TL) experiments,}
we utilize 
% Ted talks by \cite{qi2018andPre-trainedEmbd} that contains parallel data for $58$ languages aligned to English but the SATOS languages. 
\mn{the Ted talks corpus by \cite{qi2018andPre-trainedEmbd}, which contains English-aligned parallel data for $58$ languages  but  does not include the SATOS ones.} 
Monolingual data for each SATOS language are extracted from Wikipedia dumps~\footnote{Wikipedia: \url{https://dumps.wikimedia.org/}} %\mn{\textbf{[WHERE DO THESE DUMPS COME FROM? ANY REFERENCE? - Since there are multiple wiki dumping archives papers do not give specific links]}} 
and the Habit corpus~\citep{rychly2016annotatedHabit}.\footnote{Habit: \url{http://habit-project.eu/wiki/SetOfEthiopianWebCorpora}.} %We select ...
% Note that, given the data scarcity our goal is to collect all available data of the SATOS languages, and ultimately provide a standardized data, particularly multi-domain test sets for facilitating future research.
\mn{Note that, given the data scarcity conditions characterizing SATOS languages,  our goal is to collect all the  corpora available for these languages, so to ultimately provide a standardized dataset comprising multi-domain test benchmarks 
%to facilitate 
facilitating future research.}
Tables~\ref{tab:data-stat-para} and~\ref{tab:data-stat-mono} show the amount of data collected in parallel and monolingual settings.\footnote{Considering geographical location there are well over $100$ languages in the Horn of the African continent which we plan to investigate in a future work.}

At preprocessing time, data is split into train, dev and test sets. 
%To facilitate a strong evaluation criteria, we follow a random selection strategy both for the dev and test sets with an up to $5K$ segments, while the rest is left as a training data.
% To avoid potential overlap, we filter out similar segments to the dev and test sets from the different domain training data. 
\mn{To avoid bias towards specific domains, balanced dev and test sets are built by randomly selecting up to $5K$ segments. The remaining material is  left as training data, after filtering out segments similar  to those contained in the dev and test sets in order to avoid potential overlap.}
Then, the standardized data is segmented into subword units using SentencePiece~\citep{kudo2018sentencepiece}.\footnote{SentencePiece: \url{https://github.com/google/sentencepiece}} 
The segmentation rules are set to $16K$ for all models, except for the multilingual models that 
%utilizes
\mn{utilize $32K$ subwords}. When required, particularly for evaluation, the Moses toolkit~\cite{koehn2007moses} is used to tokenize/detokenize segments.
Unless otherwise specified, we use the same pre-processing stages 
%in all model training.
\mn{to train all the models.}

\subsection{Model \mn{Types} And Evaluation}
% For evaluating NMT approaches for LRL %discussed in Sec~\ref{sec:lrl-nmt-approaches},
% we train the following model types: 
\mn{To evaluate different NMT approaches on LRL\mn{s}, we train the following model types: }

\begin{enumerate}[{\it i.}]
    \item {\bf S-NMT}: a total of $10$ single language pair models trained for each {\tt SATOS} $\leftrightarrow ${\tt En} pairs.  
    
    \item {\bf SS-NMT}: semi-supervised models trained with the original parallel data of an S-NMT model and synthetic data generated with back-translation for each language pair. 
    
    \item {\bf TL}: adapted child model for each language pair parallel data from the massively multilingual parent model (M-NMT116). 
    
    \item {\bf M-NMT}: a single multilingual model trained on the aggregation of all the 
    %SATOS $\leftrightarrow$ En 
    {\tt SATOS} $\leftrightarrow ${\tt En}
    data. 
\end{enumerate}
%While we train an M-NMT model using all the available SOTAS language separately, we, however, over sample each language direction to keep an equal distribution at mini-batch sampling. 

% Models are evaluated on multi-domain test sets when available, else only the in-domain test set is used. BLEU~\citep{papineni2002bleu} is used to measure the performance of each system.\footnote{Moses Toolkit: \url{http://www.statmt.org/moses}} 
These NMT models are evaluated on multi-domain test sets when available; otherwise, only the in-domain test set is used. BLEU~\citep{papineni2002bleu} is used to measure systems' performance.\footnote{Moses Toolkit: \url{http://www.statmt.org/moses}} 
When En is the target language, BLEU scores are computed on detokenized (\textit{hypothesis}, \textit{reference}) pairs. When the target is a LRL, %we use tokenized sentences.
we report tokenized BLEU.
% Additional discussion of NMT and the model types considered for evaluation are given in Section~\ref{sec:lrl-nmt-approaches}.
\mn{Further details about the NMT model types considered in our evaluation are given in Appendix~\ref{sec:lrl-nmt-approaches}.}

\subsection{Model Settings}
% All models are trained using the  Transformer~\cite{vaswani2017attention} architecture utilizing the OpenNMT implementation~\citep{klein2017opennmt}.\footnote{OpenNMT: \url{http://opennmt.net/}} 
\mn{All the models are trained using 
the OpenNMT implementation\footnote{OpenNMT: \url{http://opennmt.net/}} \citep{klein2017opennmt} of Transformer~\citep{vaswani2017attention}}
The model parameters are set to a $512$ hidden units and embedding dimension, $4$ layers of self-attentional encoder-decoder with $8$ heads. At training time, we use $4,096$ token level batch size with a maximum sentence length of $100$. 
For inference, we keep a $32$ example level batch size, with a beam search width of $5$. LazyAdam~\citep{kingma2014adam} is applied throughout all the strategies with 
%an initial learning rate constant of $2$. 
\mn{a constant initial learning rate value of  $2$.}
%The learning rate increases linearly up to $8,000$ warm-up training steps, and decreases afterwards with an inverse square root of the training step. 
Given the sparsity of the data, dropout~\citep{srivastava2014dropout} is set to $0.3$. The multilingual models (M-NMT and M-NMT116)  are run for $1$M steps, while the S-NMT, SS-NMT, and the adaptations steps of the TL approach vary based on the amount of data used. In all runs, 
%models are observed to converge based on the validation loss.
\mn{models' convergence is checked based on the validation loss.}

\section{Results and Discussion}
Table~\ref{tab:results-1} 
%comparatively 
shows the performance of the different LRL modeling criteria with multi-domain test sets.
%Looking at the single pair NMT models (S-NMT), we observed that in all the test domains they underperformed against the SS-NMT, TL, or M-NMT models  (See the averaged (AVG) BLEU scores). 
\mn{Looking at the single pair NMT models (S-NMT), we observe that in all the test domains they underperform with respect to the SS-NMT, TL, or M-NMT models in terms of averaged (AVG) BLEU scores.}
%The only slight increase for SS-NMT is the En-Sw ({\tt47.58}BLEU) direction in the Jw300 domain. 
%
%
% Specific to each domain, the S-NMT models show to perform reasonably on the in-domain test sets. While for an out-of-domain test set (indicated with $^*$) we observe rather a large degradation.
\mn{Specifically to each domain, the S-NMT models perform reasonably well on the in-domain test sets while, for the out-of-domain Ted test set  we often observe rather large degradations.}
% For instance, for the Ted$^*$ test set of the {\tt Sw/Am/So-En} pairs, there is a consistent performance drop both in the LRL $\leftrightarrow$ En directions. 
\mn{For instance, on the {\tt Sw/Am/So-En}  Ted test sets, there is a consistent performance drop in both in the LRL $\leftrightarrow$ En translation directions.}
%
%
% The poor performance with a slight domain shift at test time shows the susceptibility of NMT in a low-resource training condition. We expect that the S-NMT model performance can be improved with the models discussed in Sec.~\ref{sec:lrl-nmt-approaches}. 
\mn{The performance drop with test sets featuring a domain shift with respect to the training data shows the susceptibility of NMT in a low-resource training condition.  We expect that the S-NMT model performance can be improved with the more robust models described in Sec.~\ref{sec:lrl-nmt-approaches}.} 

\begin{table}[!t]
    \caption{BLEU scores for the SATOS $\leftrightarrow$ En directions,
    domain-specific best performing results are {\tt highlighted} for each direction, whereas bold shows the overall best in terms of the AVG score. %\mn{\textbf{[I SEE A WARNING ON THIS TABLE (A CURLY BRACKET AT THE VERY END, BEFORE ``end center''); IT'S NOT CLEAR IF IT IS A REAL PROBLEM OR NOT...AND I'D AVOID THE RISK TO DESTROY YOUR TABLE :-) ]}}
    }
    \label{tab:results-1}
    %\small
    \begin{center}
    \resizebox{\textwidth}{!}{%
    \begin{tabular}{cl|rrrrrrrrrr}
     &
      \multicolumn{1}{c}{} &
      \multicolumn{2}{c}{{\bf S}w-En} &
      \multicolumn{2}{c}{{\bf A}m-En} &
      \multicolumn{2}{c}{{\bf T}i-En} &
      \multicolumn{2}{c}{{\bf O}m-En} &
      \multicolumn{2}{c}{{\bf S}o-En} \\  \cline{3-12} %\hline \hline
    Model &
      \multicolumn{1}{c}{Domain} &
      \multicolumn{1}{c}{En} &
      \multicolumn{1}{c}{Sw} &
      \multicolumn{1}{c}{En} &
      \multicolumn{1}{c}{Am} &
      \multicolumn{1}{c}{En} &
      \multicolumn{1}{c}{Ti} &
      \multicolumn{1}{c}{En} &
      \multicolumn{1}{c}{Om} &
      \multicolumn{1}{c}{En} &
      \multicolumn{1}{c}{So} \\ \hline \hline %\cline{1-12} \cline{2-12}
    \multirow{5}{*}{S-NMT}                        & Jw300  & 48.71 & {\tt47.58} & 32.86 & 25.72 & 29.89 & 25.54 & 26.92 & 23.38 &       &       \\
                                                & Bible  &       &       & 30.35 & 23.36 &       &       &       &       & 29.87 & 24.64 \\
                                                & Tanzil & 18.83 & 31.67 & 11.71 & 5.71  &       &       &       &       & 8.91  & 2.46  \\
                                                & Ted    & 16.63 & 11.92 & 4.26  & 1.32  &       &       &       &       & 1.35  & 0.39  \\ %\cline{3-12}
                                                & AVG    & 28.06 & 30.39 & 19.80 & 14.03 & 29.89 & 25.54 & 26.92 & 23.38 & 13.38 & 9.16  \\  \cline{2-12} 
    \multicolumn{1}{l}{\multirow{5}{*}{SS-NMT}} & Jw300  & {\tt48.90}  & 47.45 & 32.76 & {\tt26.54} & 29.84 & 25.99 & 26.45 & 23.47 &       &       \\
    \multicolumn{1}{l}{}                        & Bible  &       &       & {\tt30.53} & 24.21 &       &       &       &       & 27.68 & 22.89 \\
    \multicolumn{1}{l}{}                        & Tanzil & 19.44 & {\tt32.17} & 12.55 & 7.29  &       &       &       &       & 6.75  & 2.25  \\
    \multicolumn{1}{l}{}                        & Ted    & 18.62 & 14.72 & 6.92  & 1.41  &       &       &       &       & 1.21  & 0.52  \\ %\cline{3-12}
    \multicolumn{1}{l}{}                        & AVG    & 28.99 & {\bf31.45} & 20.69 & {\bf14.86} & 29.84 & 25.99 & 26.45 & 23.47 & 11.88 & 8.55  \\ \cline{2-12}
    \multirow{5}{*}{TL}                         & Jw300  & 48.74 & 47.39 & 32.95 & 26.49 & 29.81 & {\tt26.47} & 27.77 & 24.54 &       &       \\
                                                & Bible  &       &       & 30.36 & {\tt24.26} &       &       &       &       & 32.07 & 27.67 \\
                                                & Tanzil & {\tt19.9}  & 31.78 & 12.28 & 7.34  &       &       &       &       & 10.14 & 3.34  \\
                                                & Ted  & {\tt19.74} & {\tt14.81} & {\tt7.42}  & 1.31  &       &       &       &       & 1.97  & 0.56  \\ %\cline{3-12}
                                                & AVG    & {\bf29.46} & 31.33 & 20.75 & 14.85 & 29.81 & {\bf26.47} & 27.77 & 24.54 & 14.73 & 10.52 \\ \cline{2-12}
    \multicolumn{1}{l}{\multirow{5}{*}{M-NMT}}  & Jw300  & 46.62 & 44.47 & {\tt33.21} & 24.39 & {\tt32.21} & 26.4  & {\tt32.24} & {\tt24.96}    &       &       \\
    \multicolumn{1}{l}{}                        & Bible  &       &       & 29.78 & 20.01 &       &       &       &                              & {\tt34.99} & {\tt28.76} \\
    \multicolumn{1}{l}{}                        & Tanzil & 18.75 & 24.22 & {\tt13.68} & 10.95 &       &       &       &                         & {\tt12.68} & {\tt3.73}  \\
    \multicolumn{1}{l}{}                        & Ted    & 17.54 & 14.65 & 6.78  & 1.32  &       &       &       &                            & {\tt3.09}  & {\tt1.01}  \\ %\cline{3-12}
    \multicolumn{1}{l}{}                        & AVG    & 27.64 & 27.78 & {\bf20.86} & 14.17 & {\bf32.21} & 26.40 & {\bf32.24} & {\bf24.96}    & {\bf16.92} & {\bf11.17} \\ \hline %\hline
    \end{tabular}%
    }
    \end{center}
\end{table}

Indeed, the AVG \mn{BLEU} scores of the SS-NMT, TL, and M-NMT models show better performance in most of the cases compared to the S-NMT model.
Specifically, M-NMT 
%shows the best performance
\mn{achieves the highest results}
in six out of ten directions. Interestingly, except 
%in the
\mn{for}
 {\tt En}$\rightarrow${\tt Om/So}, all the other improvements of the M-NMT occur when translating into {\tt En}, these are: {\tt Am/Ti/Om/So}$\rightarrow${\tt En} directions. These improvements are highly related to the fact that all the LRL\mn{s} are paired with {\tt En}, maximizing the distribution of the {\tt En} data both on the encoder and decoder side. %Resulting in an M-NMT model that is robust towards the En target. 
M-NMT also shows the largest drops when compared to all the other models. These drops occur particularly for the {\tt Sw-En} pair, with a -$1.82$ and -$3.67$ 
%drop
\mn{decrease}
respectively in comparison with the best performing approaches (TL, SS-NMT). 
Similarly, a slight degradation is observed in the {\tt En}$\rightarrow${\tt Am/Ti} directions. %
%
% Our observation is that, the {\tt Sw-En} (relatively) constitutes the largest number of parallel examples, followed by {\tt Am-En} and {\tt Ti-En}. Meaning the least resourced pairs ({\tt Om-En}, {\tt So-En}) benefits the most from the M-NMT setting. 
\mn{Our observation is that  {\tt Sw-En} can exploit the  largest amount of parallel data, followed by {\tt Am-En} and {\tt Ti-En}. This indicates that the least resourced pairs ({\tt Om-En} and {\tt So-En}) benefit most from  M-NMT modeling. }
%todo: to further verify this we trained an M-NMT model without the oversampling criteria for each of the LRL pairs ... 
Moreover, it is easy to notice \mn{that} most of the performance degradation occurs when translating 
%to
\mn{into}
the LRL.

For the the SS-NMT and TL approaches, our 
%findings 
\mn{experiments}
show comparable performance in most of the translation directions. Both the approaches outperform the M-NMT in a total of four directions: SS-NMT in {\tt En}$\rightarrow${\tt  Sw/Am}, while TL in {\tt Sw}$\rightarrow${\tt  En} and {\tt En}$\rightarrow${\tt  Ti}. Contrasting 
%the 
SS-NMT and TL, the latter shows either 
%a 
comparable or better performance. 
Particularly, for the less-resourced {\tt Om/So-En} pairs, the TL approach improves over 
%the 
SS-NMT. 
Indeed, these comparisons 
%are highly dependent 
highly depend
on several factors: the type of training data (monolingual for SS-NMT, and parallel 
%data
corpora
for TL), size and data distribution, and domain mismatch between the monolingual and parallel data. For instance, for the {\tt So-En} pair, the SS-NMT shows a drop even if there is more training data from the back-translation stage. Perhaps, in addition to the 
%weak
\mn{poor quality of the}
back-translation, the drop can be attributed to the dissimilarity of the target monolingual
\mn{data} from the original parallel 
%data. 
\mn{corpora.}
%Where as the TL approach is

Moreover, domain-level performance of the SS-NMT, TL and M-NMT models shows a similar pattern as in the S-NMT. 
An interesting aspect is that the performance on the out-of-domain test set (Ted$^*$) shows a 
%better 
\mn{larger}
improvement margin than the in-domain test sets, where the best performance comes from the TL and M-NMT models. 
For instance, TL improves the {\tt Sw}$\rightarrow${\tt  En} to {\tt19.74} BLEU from the baseline S-NMT at $16.63$ and the {\tt En}$\rightarrow${\tt  Sw} to {\tt14.81} from $11.92$. 
Note that these improvements can be attributed to the domain similarity between the M-NMT116 model that is trained with Ted talks data and used for TL stages. 
%Likewise, however, 
\mn{However,}
using all the SATOS pairs, the M-NMT model improves all the out-of-domain test cases, 
%highly benefiting the extremely 
\mn{with large gains in the extremely}
low-resourced ({\tt Om/So-En}) pairs.
Overall, utilizing all the data at our disposal we can show improvements over the baseline S-NMT models. % and the conditions where SS-NMT and TL can perform poorly.
%Summary
\mn{A summary}
of open problems for LRL NMT based on the findings of this work is presented in Section~\ref{appendix:open-problems}. 

%% 
%% Open Problems - see appendix
%% 

\section{Conclusions}
In this work, we 
%showed 
\mn{analyzed}
the state of NMT approaches using five low-resource languages. Our investigation shows that the baseline 
%single pair
\mn{single-pair}
model can be 
%improved better using a 
\mn{significantly improved by the more robust}
semi-supervised, transfer-learning, and multilingual modeling approaches. However, a test on 
%an 
out-of-domain data shows \mn{the} poor performance of all the approaches. 
% We intend that this work will set the platform for the community working on the low-resource NMT modeling. 
\mn{This work will hopefully set the stage  for further research on low-resource NMT modeling. } Data, models, and scripts are available at \url{https://github.com/surafelml/Afro-NMT}. For open problems observed in this work see Section~\ref{appendix:open-problems}.

\bibliography{iclr2020_conference}

\begin{thebibliography}{42}
\providecommand{\natexlab}[1]{#1}
\providecommand{\url}[1]{\texttt{#1}}
\expandafter\ifx\csname urlstyle\endcsname\relax
  \providecommand{\doi}[1]{doi: #1}\else
  \providecommand{\doi}{doi: \begingroup \urlstyle{rm}\Url}\fi

\bibitem[Agi{\'c} \& Vuli{\'c}(2019)Agi{\'c} and
  Vuli{\'c}]{agic-vulic-2019-jw300}
{\v{Z}}eljko Agi{\'c} and Ivan Vuli{\'c}.
\newblock {JW}300: A wide-coverage parallel corpus for low-resource languages.
\newblock In \emph{Proceedings of the 57th Annual Meeting of the Association
  for Computational Linguistics}, pp.\  3204--3210, Florence, Italy, July 2019.
  Association for Computational Linguistics.
\newblock URL \url{https://www.aclweb.org/anthology/P19-1310}.

\bibitem[Arivazhagan et~al.(2019)Arivazhagan, Bapna, Firat, Lepikhin, Johnson,
  Krikun, Chen, Cao, Foster, Cherry, et~al.]{arivazhagan2019massively}
Naveen Arivazhagan, Ankur Bapna, Orhan Firat, Dmitry Lepikhin, Melvin Johnson,
  Maxim Krikun, Mia~Xu Chen, Yuan Cao, George Foster, Colin Cherry, et~al.
\newblock Massively multilingual neural machine translation in the wild:
  Findings and challenges.
\newblock \emph{arXiv preprint arXiv:1907.05019}, 2019.

\bibitem[Artetxe et~al.(2018)Artetxe, Labaka, and
  Agirre]{artetxe2018unsupervised}
Mikel Artetxe, Gorka Labaka, and Eneko Agirre.
\newblock Unsupervised statistical machine translation.
\newblock \emph{arXiv preprint arXiv:1809.01272}, 2018.

\bibitem[Bahdanau et~al.(2014)Bahdanau, Cho, and Bengio]{bahdanau2014neural}
Dzmitry Bahdanau, Kyunghyun Cho, and Yoshua Bengio.
\newblock Neural machine translation by jointly learning to align and
  translate.
\newblock \emph{arXiv preprint arXiv:1409.0473}, 2014.

\bibitem[Bentivogli et~al.(2018)Bentivogli, Bisazza, Cettolo, and
  Federico]{bentivogli:CSL2018}
Luisa Bentivogli, Arianna Bisazza, Mauro Cettolo, and Marcello Federico.
\newblock Neural versus phrase-based mt quality: An in-depth analysis on
  english-german and english-french.
\newblock \emph{Computer Speech \& Language}, 49:\penalty0 52--70, 2018.

\bibitem[Bertoldi \& Federico(2009)Bertoldi and Federico]{bertoldi2009domain}
Nicola Bertoldi and Marcello Federico.
\newblock Domain adaptation for statistical machine translation with
  monolingual resources.
\newblock In \emph{Proceedings of the fourth workshop on statistical machine
  translation}, pp.\  182--189. Association for Computational Linguistics,
  2009.

\bibitem[Caswell et~al.(2019)Caswell, Chelba, and Grangier]{caswell2019tagged}
Isaac Caswell, Ciprian Chelba, and David Grangier.
\newblock Tagged back-translation.
\newblock \emph{arXiv preprint arXiv:1906.06442}, 2019.

\bibitem[Cettolo et~al.(2012)Cettolo, Girardi, and Federico]{cettolo2012wit3}
Mauro Cettolo, Christian Girardi, and Marcello Federico.
\newblock Wit3: Web inventory of transcribed and translated talks.
\newblock In \emph{Proceedings of the 16th Conference of the European
  Association for Machine Translation (EAMT)}, volume 261, pp.\  268, 2012.

\bibitem[Cho et~al.(2014)Cho, Van~Merri{\"e}nboer, Gulcehre, Bahdanau,
  Bougares, Schwenk, and Bengio]{cho2014learningGRU}
Kyunghyun Cho, Bart Van~Merri{\"e}nboer, Caglar Gulcehre, Dzmitry Bahdanau,
  Fethi Bougares, Holger Schwenk, and Yoshua Bengio.
\newblock Learning phrase representations using rnn encoder-decoder for
  statistical machine translation.
\newblock \emph{arXiv preprint arXiv:1406.1078}, 2014.

\bibitem[Christodouloupoulos \& Steedman(2015)Christodouloupoulos and
  Steedman]{christodouloupoulos2015massively}
Christos Christodouloupoulos and Mark Steedman.
\newblock A massively parallel corpus: the bible in 100 languages.
\newblock \emph{Language resources and evaluation}, 49\penalty0 (2):\penalty0
  375--395, 2015.

\bibitem[Dong et~al.(2015)Dong, Wu, He, Yu, and Wang]{dong2015multi}
Daxiang Dong, Hua Wu, Wei He, Dianhai Yu, and Haifeng Wang.
\newblock Multi-task learning for multiple language translation.
\newblock In \emph{ACL (1)}, pp.\  1723--1732, 2015.

\bibitem[Edunov et~al.(2018)Edunov, Ott, Auli, and
  Grangier]{edunov2018understanding}
Sergey Edunov, Myle Ott, Michael Auli, and David Grangier.
\newblock Understanding back-translation at scale.
\newblock \emph{arXiv preprint arXiv:1808.09381}, 2018.

\bibitem[Firat et~al.(2016)Firat, Cho, and Bengio]{firat2016multi}
Orhan Firat, Kyunghyun Cho, and Yoshua Bengio.
\newblock Multi-way, multilingual neural machine translation with a shared
  attention mechanism.
\newblock \emph{arXiv preprint arXiv:1601.01073}, 2016.

\bibitem[Gehring et~al.(2017)Gehring, Auli, Grangier, Yarats, and
  Dauphin]{gehring2017convolutional}
Jonas Gehring, Michael Auli, David Grangier, Denis Yarats, and Yann~N Dauphin.
\newblock Convolutional sequence to sequence learning.
\newblock In \emph{Proceedings of the 34th International Conference on Machine
  Learning-Volume 70}, pp.\  1243--1252. JMLR. org, 2017.

\bibitem[Gu et~al.(2019)Gu, Wang, Cho, and Li]{gu2019improved}
Jiatao Gu, Yong Wang, Kyunghyun Cho, and Victor~OK Li.
\newblock Improved zero-shot neural machine translation via ignoring spurious
  correlations.
\newblock \emph{arXiv preprint arXiv:1906.01181}, 2019.

\bibitem[Guzm{\'a}n et~al.(2019)Guzm{\'a}n, Chen, Ott, Pino, Lample, Koehn,
  Chaudhary, and Ranzato]{guzman2019two}
Francisco Guzm{\'a}n, Peng-Jen Chen, Myle Ott, Juan Pino, Guillaume Lample,
  Philipp Koehn, Vishrav Chaudhary, and Marc'Aurelio Ranzato.
\newblock Two new evaluation datasets for low-resource machine translation:
  Nepali-english and sinhala-english.
\newblock \emph{arXiv preprint arXiv:1902.01382}, 2019.

\bibitem[Ha et~al.(2016)Ha, Niehues, and Waibel]{ha2016toward}
Thanh-Le Ha, Jan Niehues, and Alexander Waibel.
\newblock Toward multilingual neural machine translation with universal encoder
  and decoder.
\newblock \emph{arXiv preprint arXiv:1611.04798}, 2016.

\bibitem[Johnson et~al.(2017)Johnson, Schuster, Le, Krikun, Wu, Chen, Thorat,
  Vi{\'e}gas, Wattenberg, Corrado, Hughes, and Dean]{johnson2016google}
Melvin Johnson, Mike Schuster, Quoc~V Le, Maxim Krikun, Yonghui Wu, Zhifeng
  Chen, Nikhil Thorat, Fernanda Vi{\'e}gas, Martin Wattenberg, Greg Corrado,
  Macduff Hughes, and Jeffrey Dean.
\newblock Google's multilingual neural machine translation system: Enabling
  zero-shot translation.
\newblock \emph{Transactions of the Association for Computational Linguistic},
  5:\penalty0 339--351, 2017.
\newblock URL \url{https://aclweb.org/anthology/Q/Q17/Q17-1024.pdf}.

\bibitem[Kalchbrenner \& Blunsom(2013)Kalchbrenner and
  Blunsom]{kalchbrenner2013recurrent}
Nal Kalchbrenner and Phil Blunsom.
\newblock Recurrent continuous translation models.
\newblock In \emph{Proceedings of the 2013 Conference on Empirical Methods in
  Natural Language Processing}, pp.\  1700--1709, 2013.

\bibitem[Kingma \& Ba(2014)Kingma and Ba]{kingma2014adam}
Diederik Kingma and Jimmy Ba.
\newblock Adam: A method for stochastic optimization.
\newblock \emph{arXiv preprint arXiv:1412.6980}, 2014.

\bibitem[Klein et~al.(2017)Klein, Kim, Deng, Senellart, and
  Rush]{klein2017opennmt}
Guillaume Klein, Yoon Kim, Yuntian Deng, Jean Senellart, and Alexander~M Rush.
\newblock Opennmt: Open-source toolkit for neural machine translation.
\newblock \emph{arXiv preprint arXiv:1701.02810}, 2017.

\bibitem[Kocmi \& Bojar(2018)Kocmi and Bojar]{kocmi2018trivial}
Tom Kocmi and Ond{\v{r}}ej Bojar.
\newblock Trivial transfer learning for low-resource neural machine
  translation.
\newblock \emph{arXiv preprint arXiv:1809.00357}, 2018.

\bibitem[Koehn \& Knowles(2017)Koehn and Knowles]{koehn2017six}
Philipp Koehn and Rebecca Knowles.
\newblock Six challenges for neural machine translation.
\newblock \emph{arXiv preprint arXiv:1706.03872}, 2017.

\bibitem[Koehn et~al.(2007)Koehn, Hoang, Birch, Callison-Burch,
  et~al.]{koehn2007moses}
Philipp Koehn, Hieu Hoang, Alexandra Birch, Chris Callison-Burch, et~al.
\newblock Moses: Open source toolkit for statistical machine translation.
\newblock In \emph{Proc. of ACL}, 2007.

\bibitem[Kudo \& Richardson(2018)Kudo and Richardson]{kudo2018sentencepiece}
Taku Kudo and John Richardson.
\newblock Sentencepiece: A simple and language independent subword tokenizer
  and detokenizer for neural text processing.
\newblock \emph{arXiv preprint arXiv:1808.06226}, 2018.

\bibitem[Lakew et~al.(2017{\natexlab{a}})Lakew, Lotito, Matteo, Marco, and
  Marcello]{lakew2017improving}
Surafel~M Lakew, Quintino~F Lotito, Negri Matteo, Turchi Marco, and Federico
  Marcello.
\newblock Improving zero-shot translation of low-resource languages.
\newblock In \emph{14th International Workshop on Spoken Language Translation},
  2017{\natexlab{a}}.

\bibitem[Lakew et~al.(2017{\natexlab{b}})Lakew, Di~Gangi, and
  Federico]{lakew17mnmt-lrl}
Surafel~Melaku Lakew, Mattia~Antonino Di~Gangi, and Marcello Federico.
\newblock {``Multilingual Neural Machine Translation for Low Resource
  Languages''}.
\newblock In \emph{Proceedings of the 4th Italian Conference on Computational
  Linguistics (CLiC-IT)}, Rome, Italy, 2017{\natexlab{b}}.

\bibitem[Lakew et~al.(2018)Lakew, Erofeeva, Negri, Federico, and
  Turchi]{lakew18:tl-dv}
Surafel~Melaku Lakew, Aliia Erofeeva, Matteo Negri, Marcello Federico, and
  Marco Turchi.
\newblock {``Transfer Learning in Multilingual Neural Machine Translation with
  Dynamic Vocabulary''}.
\newblock In \emph{15th International Workshop on Spoken Language Translation
  (IWSLT)}, Bruges, Belgium, 2018.

\bibitem[Lample et~al.(2018)Lample, Denoyer, and
  Ranzato]{lample2017unsupervised}
Guillaume Lample, Ludovic Denoyer, and Marc'Aurelio Ranzato.
\newblock Unsupervised machine translation using monolingual corpora only.
\newblock In \emph{Proceedings of the 6th International Conference on Learning
  Representations}, 2018.

\bibitem[Neubig \& Hu(2018)Neubig and Hu]{neubig18:rapid}
Graham Neubig and Junjie Hu.
\newblock Rapid adaptation of neural machine translation to new languages.
\newblock In \emph{Proceedings of the 2018 Conference on Empirical Methods in
  Natural Language Processing}, pp.\  875--880. Association for Computational
  Linguistics, 2018.
\newblock URL \url{https://aclweb.org/anthology/D18-1103}.

\bibitem[Nguyen \& Chiang(2017)Nguyen and Chiang]{nguyen2017transfer}
Toan~Q Nguyen and David Chiang.
\newblock Transfer learning across low-resource, related languages for neural
  machine translation.
\newblock \emph{arXiv preprint arXiv:1708.09803}, 2017.

\bibitem[Papineni et~al.(2002)Papineni, Roukos, Ward, and
  Zhu]{papineni2002bleu}
Kishore Papineni, Salim Roukos, Todd Ward, and Wei-Jing Zhu.
\newblock Bleu: a method for automatic evaluation of machine translation.
\newblock In \emph{Proceedings of the 40th annual meeting on association for
  computational linguistics}, pp.\  311--318. Association for Computational
  Linguistics, 2002.

\bibitem[Qi et~al.(2018)Qi, Sachan, Felix, Padmanabhan, and
  Neubig]{qi2018andPre-trainedEmbd}
Ye~Qi, Devendra~Singh Sachan, Matthieu Felix, Sarguna~Janani Padmanabhan, and
  Graham Neubig.
\newblock When and why are pre-trained word embeddings useful for neural
  machine translation?
\newblock In \emph{Proceedings of NAACL-HLT 2018}, pp.\  529--535. Association
  for Computational Linguistics, 2018.
\newblock URL \url{https://aclweb.org/anthology/N18-2084}.

\bibitem[Rychl{\`y} \& Suchomel(2016)Rychl{\`y} and
  Suchomel]{rychly2016annotatedHabit}
Pavel Rychl{\`y} and V{\'\i}t Suchomel.
\newblock Annotated amharic corpora.
\newblock In \emph{International Conference on Text, Speech, and Dialogue},
  pp.\  295--302. Springer, 2016.

\bibitem[Sennrich et~al.(2015)Sennrich, Haddow, and
  Birch]{sennrich2015improvingMono}
Rico Sennrich, Barry Haddow, and Alexandra Birch.
\newblock Improving neural machine translation models with monolingual data.
\newblock \emph{arXiv preprint arXiv:1511.06709}, 2015.

\bibitem[Sennrich et~al.(2016)Sennrich, Haddow, and Birch]{sennrich15:subword}
Rico Sennrich, Barry Haddow, and Alexandra Birch.
\newblock Neural machine translation of rare words with subword units.
\newblock In \emph{Proceedings of the 54th Annual Meeting of the Association
  for Computational Linguistics}, pp.\  1715--1725. Association for
  Computational Linguistics (ACL), 8 2016.
\newblock ISBN 978-1-945626-03-6.
\newblock \doi{10.18653/v1/P16-1162}.

\bibitem[Srivastava et~al.(2014)Srivastava, Hinton, Krizhevsky, Sutskever, and
  Salakhutdinov]{srivastava2014dropout}
Nitish Srivastava, Geoffrey~E Hinton, Alex Krizhevsky, Ilya Sutskever, and
  Ruslan Salakhutdinov.
\newblock Dropout: a simple way to prevent neural networks from overfitting.
\newblock \emph{Journal of machine learning research}, 15\penalty0
  (1):\penalty0 1929--1958, 2014.

\bibitem[Sutskever et~al.(2014)Sutskever, Vinyals, and
  Le]{sutskever2014sequence}
Ilya Sutskever, Oriol Vinyals, and Quoc~V Le.
\newblock Sequence to sequence learning with neural networks.
\newblock In \emph{Advances in neural information processing systems}, pp.\
  3104--3112, 2014.

\bibitem[Tachbelie et~al.(2009)Tachbelie, Abate, and
  Menzel]{tachbelie2009morphemeAmharic}
Martha~Yifiru Tachbelie, Solomon~Teferra Abate, and Wolfgang Menzel.
\newblock Morpheme-based language modeling for amharic speech recognition.
\newblock In \emph{Human Language Technology. Challenges for Computer Science
  and Linguistics.} Citeseer, 2009.

\bibitem[Tiedemann(2012)]{tiedemann2012parallel}
J{\"o}rg Tiedemann.
\newblock Parallel data, tools and interfaces in opus.
\newblock In \emph{Proceedings of Language Resources and Evaluation (LREC)},
  2012.

\bibitem[Vaswani et~al.(2017)Vaswani, Shazeer, Parmar, Uszkoreit, Jones, Gomez,
  Kaiser, and Polosukhin]{vaswani2017attention}
Ashish Vaswani, Noam Shazeer, Niki Parmar, Jakob Uszkoreit, Llion Jones,
  Aidan~N Gomez, {\L}ukasz Kaiser, and Illia Polosukhin.
\newblock Attention is all you need.
\newblock In \emph{Advances in Neural Information Processing Systems}, pp.\
  6000--6010, 2017.

\bibitem[Zoph et~al.(2016)Zoph, Yuret, May, and Knight]{zoph16:tf}
Barret Zoph, Deniz Yuret, Jonathan May, and Kevin Knight.
\newblock Transfer learning for low-resource neural machine translation.
\newblock In \emph{Proceedings of the 2016 Conference on Empirical Methods in
  Natural Language Processing}, pp.\  1568--1575. Association for Computational
  Linguistics, 2016.
\newblock URL \url{https://aclweb.org/anthology/D16-1163}.

\end{thebibliography}
\bibliographystyle{iclr2020_conference}

\newpage
\appendix
\section{Appendix}

\subsection{Low-Resource (SATOS) Languages} %SATOS Languages
\begin{table}[!ht]%todo: insert fig of horn-south showing the coverage of these languages on the right. 
\centering
    \caption{Background on the SATOS languages that are considered in this work. Number of speakers is following the estimates provided by~\url{https://www.ethnologue.com} (2015).}\label{tab:lang-stat} \vspace{1em}
    \begin{tabular}{llllr}
                    & \multicolumn{4}{c}{Description} \\ \cline{2-5} \\ %\hline \\
    Language        & Family        & Sub-Family    & Script    & Users/Speakers \\ \hline \hline
    Swahili (Sw)         & Niger-Congo  & Bantu         & Latin     & ~150M \\
    Amharic (Am)        & Afroasiatic  & South-Semitic & Ge'ez/Ethiopic    & ~25M \\
    Tigrigna (Ti)        & Afroasiatic  & South-Semitic & Ge'ez/Ethiopic    & ~7M \\
    Oromo (Om)           & Afroasiatic  & Chustic        & Latin             & ~35M\\
    Somali (So)         & Afroasiatic  & Chustic        & Latin             & ~16M\\ \hline
    %English         & Indo-European & Germanic      & Latin             & ~600M \\ \hline
    \end{tabular}
\end{table}

\subsection{Data and Statistics}
\begin{table}[!ht]
\centering
\caption{Data statistics in number of examples for each pair of the SATOS languages paired with English, across four domains.} 
\label{tab:data-stat-para} \vspace{1em}
%\resizebox{\textwidth}{!}{%
\begin{tabular}{cllllll}
\multicolumn{1}{l}{}              &       & \multicolumn{5}{c}{Domain}              \\ \cline{2-7} \\
\multicolumn{1}{l}{Language Pair} & Split & Jw300  & Bible & Tanzil & Ted  & Total  \\ \hline \hline
\multirow{3}{*}{Sw-En}            & train & 907842 &       & 87645  &      & 995487 \\
                                  & dev   & 5179   &       & 3505   & 681  & 9365   \\
                                  & test  & 5315   &       & 3509   & 1364 & 10188  \\
\multirow{3}{*}{Am-En}            & train & 538677 & 43172 & 17461  &      & 599310 \\
                                  & dev   & 4514   & 4685  & 4905   &      & 14104  \\
                                  & test  & 4551   & 4685  & 4911   & 567  & 14714  \\
\multirow{3}{*}{Ti-En}            & train & 344540 &       &        &      & 344540 \\
                                  & dev   & 4845   &       &        &      & 4845   \\
                                  & test  & 4945   &       &        &      & 4945   \\
\multirow{3}{*}{Om-En}            & train & 907842 &       &        &      & 907842 \\
                                  & dev   & 5179   &       &        &      & 5179   \\
                                  & test  & 5315   &       &        &      & 5315   \\
\multirow{3}{*}{So-En}            & train &        & 44276 & 24592  &      & 68868  \\
                                  & dev   &        & 4713  & 4393   & 565  & 9671   \\
                                  & test  &        & 4735  & 4450   & 1132 & 10317 \\ \hline
\end{tabular}%
%}
\end{table}

\begin{table}[!ht]
    \centering
    \caption{Monolingual data size of the SATOS languages collected from Wiki dump and the Habit~\citep{rychly2016annotatedHabit} project. Note, the En side monolingual is only from Wiki and selected proportional with each LRL monolingual data.} \label{tab:data-stat-mono} \vspace{1em}
    \begin{tabular}{llllll}
          & \multicolumn{4}{c}{Language} \\ \cline{2-6} %\\
          &  Sw    & Am     & Ti      & Om     & So    \\ \hline \hline %\cline{2-6}
    Wiki  & 351805 & 114251  & 2560   & 12162  & 69386   \\
    Habit &        & 1208947 & 139357 & 250432 & 2643337 \\ 
    Total & 351805 & 1323198 & 141917 & 262594 & 2712723 \\ \hline \hline
    \end{tabular}
\end{table}

\subsection{NMT Approaches for Low-Resource Languages}% \label{appendix:nmt-for-lrl}
\label{sec:lrl-nmt-approaches}
\subsubsection{Neural Machine Translation}
MT is a task of mapping a source language sequence $X=x_1,x_2,\ldots,x_{L_x}$ into a target language $Y=y_1,y_2\ldots,y_{L_Y}$, where $L_x$ and $L_y$ can differ.
Several types of architectures have been proposed for modeling NMT: Recurrent~\citep{kalchbrenner2013recurrent,sutskever2014sequence,cho2014learningGRU,bahdanau2014neural}, Convolutional~\citep{gehring2017convolutional}, and recently Transformer (TNN) by~\cite{vaswani2017attention} that have shown better performance and efficient processing of input tokens in a simultaneous manner.
Though there are different formalizations of NMT for sequence representation, the common underlying principle is to learn a model in an end-to-end fashion. % by back-propagating gradients from the decoder network all the way back to the encoder.
In general, an {\it encoder} network reads the input sequence ($X$) and creates a latent representation of it, whereas a {\it decoder} network learns how to generate the output sequence ($Y$). 
In this work, we utilize the TNN for modeling the NMT systems.

TNN is built using a mechanism called {\it self-attention}, that computes relations between the different positions of a given sequence to generate hidden representations. %Partly analogous to the bidrectional RNN, 
Both the encoder and decoder of TNN constitute a stack of self-attention layers followed by a fully-connected feed-forward (FNN) layers.
The encoder is composed of $N$ number of similar layers. Each of the $N$ layers comprises two sub-layers. The first sub-layer is a multi-headed self-attention, while the second is an FNN. 
The decoder side is similar to the encoder, except a third multi-head self-attention layer is added, to specifically attend on the encoder representation.
%There are several strategies to implement a decoder but all of them end up computing the conditional probability of the next target word depending on the previously translated words and the source sentence:
For each target token prediction stage, a conditional probability is computed using the previously decoded token and the source sequence ($X$):

\begin{equation}
    p(y_i = k\vert y_{<i}, \mathbf{x})
\end{equation}

The network is trained end-to-end to find the parameters $\mathbf{\hat{\theta}}$ that maximizes the log-likelihood of the training set $\{(\mathbf{x}_t,\mathbf{y}_t): t=1,\ldots,L_y\}$ :
%\[\sum_{s=1}^{S} \log p(\mathbf{e}_s\vert \mathbf{f}_s;\Theta) \]
\begin{equation}
    \sum_{t=1}^{L_y} \log p(\mathbf{y}_t\vert \mathbf{x}_t;{\hat\theta})
\end{equation}
%}

\begin{figure}[!t]
    \caption{NMT modeling, from left to right: single pair, semi-supervised, multilingual, and transfer-learning strategies.}
    \begin{center}
        \includegraphics[width=1\textwidth]{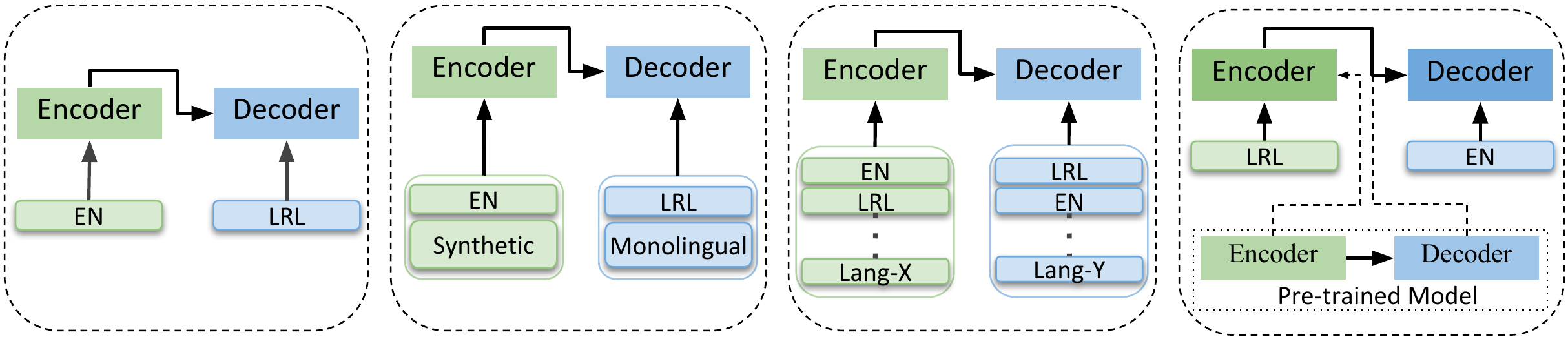}
    \end{center}
    \label{fig:nmt-modeling-types}
\end{figure}

The standard NMT (S-NMT) training requires the availability of a source to target language aligned parallel corpus.
Hence, the objective function is simply to learn the mapping from the source %($S$) 
and target %($T$) 
training examples. %In addition to a single language pair training examples, 
Moreover, several training objectives have been suggested for NMT training, as illustrated in Figure~\ref{fig:nmt-modeling-types}, which we will discuss in the following sections. %below. % these are: % with scarcity of training examples several 

\subsubsection{Semi-Supervised NMT}
In semi-supervised NMT (SS-NMT) monolingual data is utilized to improve over the S-NMT model. The primary way of achieving SS-NMT is known as {\it back-translation}~\citep{bertoldi2009domain,sennrich2015improvingMono}. Hence, to improve a Source $\rightarrow$ Target model with target language monolingual data, an SS-NMT can be formalized in three stages: {\it i}) train Target $\rightarrow$ Source model by reversing the parallel data, {\it ii}) translate the target monolingual data with the reverse model, and {\it iii}) train the Source $\rightarrow$ Target model by merging the original and the newly generated synthetic parallel data.

The expectation is that, with the augmented data, the Source $\rightarrow$ Target translation performance can be further improved. There are other variants of back-translation based SS-NMT~\citep{edunov2018understanding,caswell2019tagged}, however, for this work we focus on the above three stages following~\cite{sennrich2015improvingMono}. 
%Recently, more variants of back-translation have been proposed. Adding noise with a certain probability in the synthetic source~\cite{edunov2018understanding}, alternatively, in~\cite{caswell2019tagged} a tagged synthetic source as in multilingual NMT~\cite{johnson2016google} have shown better improvements.

\subsubsection{Transfer-Learning Based NMT}
\cite{zoph16:tf} proposed a TL paradigm, where a model trained with a high-resource pair ({\it parent}) is used to initialize a model training for LRL pair ({\it child}). Later the TL approach is improved by incorporating related languages in the parent-child transfer setup~\citep{nguyen2017transfer,kocmi2018trivial}. The parent can also be trained with a large scale multilingual data~\citep{neubig18:rapid} and adapted to the LRL pair. Moreover, by tailoring the parent vocabulary and associated model parameters %(embeddings, pre soft-max projection layer) 
to the child/new LRL pair~\citep{lakew17mnmt-lrl} have shown a better positive TL, also known as {\it dynamic TL}.

Given the diversity of languages and writing scripts, in this work, we utilize the dynamic TL mechanism following the experimental setup in~\cite{lakew18:tl-dv}. In other words, assuming a parent model pre-trained with large scale multilingual data, but not the SATOS languages, the TL stage must involve the customization to the LRL pair. Our goal is to investigate how far the pre-trained model helps to improve a new LRL pair than comparing the different TL approaches. %, despite a possible domain shift.

\subsubsection{Multilingual NMT}
M-NMT can be considered under the umbrella of TL approach, however, within a single (parent) model that aggregates all the parallel data of $N$ language pairs. 
Hence, the TL can occur implicitly, based on the assumption that the combination of all the available pairs data brings more diversity to the model training corpus. 
Though there are several M-NMT modeling  mechanisms~\citep{dong2015multi,firat2016multi}, we follow the single encoder-decoder based approach~\citep{johnson2016google,ha2016toward}, that works by appending target {\it language-flag} at the beginning of each source language example. Our goal is to comparatively evaluate the significance of the M-NMT model that leverages the aggregation of all the SATOS languages data.

\subsection{Open Problems}\label{appendix:open-problems}
The reported results in Table~\ref{tab:results-1}, and the discussion confirms what has been reported in the literature on back-translation, transfer-learning, and multilingual modeling to improve LRL translation tasks. However, there are still open problems that require further investigation with respect to the SATOS languages and other languages with small training data:  %Below, we identify and summarize the key challenges in LRL NMT modeling. 

{\bf Language and Data:}
As shown in Table~\ref{tab:lang-stat}, we explored five languages that are low-resource, as well as highly diverse. Where the varied characteristics of these languages can pose new challenges, more so in the low-resource NMT setting.  
Meaning, each language can exhibit its characteristics that might require a specialized modeling criterion. For instance, Am and Ti is a highly morphological language~\citep{tachbelie2009morphemeAmharic}, that might be improved with alternative input modeling methods than the segmentation approach~\citep{kudo2018sentencepiece} we utilized in this work. 
More importantly the availability of model training resources both in parallel and monolingual format is limited. Hence, data generation approaches that can diversify the existing examples can be a key ingredient to further improve the current model performance. In this direction,~\cite{arivazhagan2019massively} indicated the importance of formulating sample efficient learning algorithms and approaches that can leverage from other forms of data, such as speech and images.

{\bf Domain Shift:} can be characterized by scenarios such as domain imbalance within a training data or the domain mismatch between a parallel and monolingual data.   
The poor performance of each modeling type on the Ted talks are a good indication to easily identify and assess the weakness of NMT, more so in the low-resource setting. Moreover, the poor performance of the SS-NMT is another example where back-translation can also harm the initial model (S-NMT) performance if the monolingual data is too distant from the in-domain data. 
Thus, with the absence of enough training material, learning a better translation model by exploiting all available domains is an important criterion. This direction requires a model that can generalize well across domains while minimizing the negative effect as observed in the SS-NMT case.

{\bf Zero-Resource Language:} 
As we have noted in Sec~\ref{sec:intro}, the majority of the world languages do not have parallel training material. Hence, for languages pairs with only monolingual data (i.e., {\it zero-resource languages}), alternative modeling strategies are needed. We highlight this aspect, considering a real low-resource NMT modeling should aim at enabling and improving translation between the LRL pairs. 
Indeed, recent progress in zero-shot~\cite{johnson2016google,ha2016toward} %,lakew2017improving} 
and unsupervised~\cite{artetxe2018unsupervised,lample2017unsupervised} approaches remain as the primary options to explore. 
However, in light of recent studies~\citep{neubig18:rapid,guzman2019two}, that shows the weakness of zero-resource approaches, further investigation are required for languages such as SATOS.  
In other words, certain LRL share few similarities (e.g. {\it Am Vs. Sw}), and with the absence of comparable and large amount of monolingual data, zero-resource NMT settings become highly challenging. %In this regard, an interesting question could be learning a usable translation model between the LRL pairs ({\tt SATOS $\leftrightarrow$ SATOS}).
In such type of resource scarce setting, perhaps incrementally learning and improving zero-resource directions from monolingual data by leveraging multilingual model~\citep{lakew2017improving,gu2019improved} could be a promising alternative to investigate.

\end{document}